# BEST-KAG: Enhancing Question Answering of Building Engineering Standards with Multimodal Knowledge Graph Modeling and Large Language Model

Jia-Rui Lin [a,b], Junxi Guo [a,b,], Keyin Chen [a,b], Peng Pan [a,b,*]

[a] Department of Civil Engineering, Tsinghua University, Beijing, China, 100084

[b] Key Laboratory of Digital Construction and Digital Twin, Ministry of Housing and Urban-Rural Development, Beijing, China, 100084

## Abstract

Construction standards are critical for building safety and sustainability. Existing standard application workflows rely on keyword-based document retrieval and manual cross-clause interpretation, which cannot reliably support multi-clause reasoning, multimodal knowledge utilization, or traceable clause-level evidence linkage. To address these limitations, this study develops a multimodal knowledge-driven framework that supports question answering on standard knowledge named BEST-KAG (Knowledge-Augmented Generation for Building Engineering STandards). The framework introduces (i) a multimodal knowledge graph (MKG) for unified representation of document hierarchy and heterogeneous standard knowledge with various connections, (ii) a rule–LLM hybrid knowledge construction pipeline for scalable multimodal knowledge extraction, creating a large MAG with 251 building engineering standards, 171,652 nodes and 310,914 edges, and (iii) a graph-retrieval-based knowledge-augmented generation architecture for clause-grounded and traceable question answering. Experiments demonstrate that BEST-KAG consistently outperforms multiple mainstream LLMs in terms of Expert evaluation, and metrics including BLEU, and ROUGE, with the best improvement up to 74.01% compared to the baselines.

## 1 Introduction

Building engineering standards are essential instruments for governing design, construction, and operation, and they underpin safety, sustainability, and the overall quality of the built environment[1]. As digital construction technologies and data-rich workflows advance, practitioners increasingly expect standards to be not only readable documents but also computable resources that can support fast, accurate, and explainable decision-making in engineering tasks and automated checking pipelines. This demand is amplified by the growing scale and heterogeneity of digital construction data and by the need to integrate regulatory knowledge with downstream digital processes. Accordingly, improving how standards are represented, accessed, and operationalized has become a key part of the broader agenda of construction digitalization and intelligent applications[2].

When people realized that reading the standard documents directly was extremely inefficient, they began to think about how to conduct searches based on the documents in order to obtain the answers to their questions[3]. In practice, standard document retrieval over standards has long been dominated by document-centric and keyword-driven workflows. Engineers typically search PDFs or repositories using keywords, then manually locate relevant clauses, interpret cross-references, reconcile inconsistencies across documents, and finally translate textual requirements into actionable design or checking steps. This pipeline is inherently brittle: when a query does not match the original wording, retrieval becomes noisy; when a requirement is distributed across multiple clauses, tables, or formulas, manual cross-navigation is slow and error-prone; and when an answer requires procedural reasoning or numerical constraints, purely text-based lookup provides limited support[4]. Early digitalization efforts have improved storage and distribution, but they rarely change the core interaction paradigm: users still "search text" rather than "ask questions and obtain grounded answers with evidence." As a result, the industry continues to lack a robust, scalable, and traceable QA mechanism that can faithfully connect user questions to authoritative clauses and multimodal evidence[5].

With the rise of large language models, recent studies have increasingly leveraged their question-answering and reasoning capabilities to address the inefficiency and low precision of traditional standard document retrieval. Existing work typically combines LLMs with retrieval-based frameworks, such as Retrieval-Augmented Generation (RAG), to ground model responses in relevant clauses, enabling more accurate and traceable answers at the clause level [4]. Some approaches further integrate domain knowledge bases or knowledge graph (KG) through prompt engineering or hybrid reasoning mechanisms, allowing LLMs to align natural language queries with structured regulatory knowledge [5]. These methods shift standard search from keyword matching toward semantically driven, knowledge-supported question answering.

Existing research demonstrates promising attempts to model and apply knowledge derived from regulatory documents; however, three significant gaps remain [6]. First, a unified digital model is lacking. No machine-readable structure applicable across all building standards currently exists. Although PDF and XML formats facilitate storage, they fail to capture deeper logical relationships within standards. Existing ontology-based studies tend to focus on specific domains or narrow document sets and thus lack generalizability [7]. Second, multimodal knowledge extraction remains insufficient. Large-scale knowledge-base construction still depends heavily on manual expert annotation, which is labor-intensive and prone to errors. While some studies attempt automatic semantic similarity mining at the clause level, these focus primarily on text and do not cover formulas, tables, or other non-textual content. Third, intelligent applications remain limited. Manual search continues to dominate standard utilization. Even frameworks adopting Retrieval-Augmented Generation (RAG) lack domain-specific optimization, preventing deeper mining of embedded engineering knowledge[8].

To address these issues, this study proposes BEST-KAG, a unified digital and intelligent framework for building engineering standards that integrates multimodal knowledge graph-based storage, LLM-assisted knowledge extraction, and knowledge-augmented generation. The framework enables clause-level semantic retrieval and traceable, multimodal-grounded answers, thereby improving the reliability and efficiency of standard search and interpretation.

The remainder of this paper is organized as follows. Section 2 reviews related research on standard modeling, knowledge extraction, and intelligent applications. Section 3 introduces the technical details of the proposed methodology. Section 4 presents experimental results and discussion. Section 5 concludes the study and outlines future research directions.

## 2. Related Work

### 2.1 Knowledge Modeling and Representation for Standards

The architecture of a standard knowledge model directly influences the scope and mechanisms of information retrieval in practical applications. Effective modeling of standard knowledge serves as the foundation for compliance checking [4]. In the Architecture, Engineering, and Construction (AEC) domain, the rapid expansion of digital technologies has generated increasingly large volumes of data, creating an need for methods that can extract and represent knowledge from heterogeneous sources. Against this background, ontologies have gradually been adopted for organizing and formalizing knowledge embedded within regulatory standards [6].

Common ontology representation languages include RDF(S), OWL. RDF(S) is often used as a general-purpose data serialization format, while OWL is favored for its reasoning capabilitie [7]. Zhang et al. [8], for example, developed a domain ontology for the building sector that includes a concept taxonomy, definitions of relationship types, rule constraints, and ontology-based modeling of semantic elements specific to building regulations,supporting rule-based compliance checking under predefined ontology structures. Xu et al. [9] designed two ontologies for underground utility regulations, enabling the formal representation of cross-domain terminology and spatial language, thereby supporting semantic information extraction. However, such ontology studies generally focus on narrow semantic layers and application-specific domains, limiting their scalability and their ability to cover large volumes of standard documents.

Knowledge graphs are used for organizing large-scale regulatory data and relationship retrieval, but typically lack explicit modeling of clause dependencies across standards [10]. A knowledge graph typically consists of an ontology schema layer and a data layer, where fragmented information is integrated into a highly connected structured form [11]. This approach has been widely applied to the aggregation and storage of standard knowledge. Jiang et al. [12] pioneered the use of knowledge graph technology for construction safety standards, creating a structured knowledge system and corresponding intelligent retrieval applications. Liu et al. [13] extracted safety knowledge from highway engineering standards to construct a knowledge graph that supports intelligent querying; however, their work did not account for the hierarchical organization of standard clauses, which may lead to the loss or fragmentation of semantic completeness. Liu et al. [14] proposed a two-dimensional knowledge graph framework containing both clause hierarchy and clause-concept hierarchy for shield tunneling construction standards. Although effective for multi-granularity knowledge organization, the model was validated only on a small set of metro tunneling standards, and its generalizability to the broader AEC domain remains to be established. To date, the field still lacks a unified, digital and multimodal model capable of comprehensively representing building engineering standards.

**2.2 Multimodal knowledge extraction and storage**

Knowledge extraction transforms large-scale heterogeneous and multimodal data into ontology-aligned semantic representations. Its objective is to transform source data into machine-interpretable knowledge for retrieval and reasoning tasks. [15]. This process must accommodate diverse forms of content-including text, tables, and mathematical expressions-to ensure accuracy and completeness. Named Entity Recognition (NER) and Relation Extraction (RE) constitute the two essential tasks in knowledge extraction. NER focuses on identifying predefined categories of entities such as key elements, attributes, and attribute values from unstructured text, while RE mines semantic associations among these entities-for example, comparative relations or constraint relationships. Extracted entities and relations are typically represented as structured triples [16], forming the foundation of the knowledge graph.

Traditional NER approaches generally fall into two categories: rule-based and machine learning-based. Rule-based methods-such as lexicon construction and pattern matching-are simple, interpretable, and domain-adaptable. For instance, Ren et al. [17] developed a domain lexicon for construction to identify operational actions and non-physical entities, using part-of-speech tagging to support NER. However, rule construction is labor-intensive and scales poorly. Machine learning-based approaches leverage trained models to identify entities from text, offering stronger generalization for large-scale data. Hu et al. [18] implemented Chinese NER using a conditional random field (CRF) model at both character and word levels. Zhang et al. [19] proposed a maximum-entropy-based Chinese NER model that integrates heuristic rules and statistical features. Yet machine learning approaches may suffer from error propagation during feature extraction [20]. Relation extraction shares similar methodological foundations with NER, relying on rule-based systems, machine learning, or deep learning techniques.

With the rapid advancement of deep learning, large language models (LLMs) have demonstrated capabilities in text generation, few-shot and zero-shot learning, and logical reasoning.

This has opened new opportunities for NER and RE tasks [21]. Zheng et al. [22] developed a construction-domain LLM that outperformed general-purpose models (e.g., Google, Baidu) on NER tasks. Zhou et al. [23] fine-tuned the BERT model using limited labeled data to effectively annotate semantic information in building standards. Feng et al. [24] combined a BiLSTM-CRF architecture with a combinatorial data augmentation strategy to improve robustness under limited supervision. Wang et al. [25] applied prompt engineering to enable ChatGPT-3 to perform NER with strong performance under resource-constrained settings. To address efficiency bottlenecks in applying LLMs to sequence labeling tasks, Liu et al. [26] proposed an accelerated method that supports large-scale deployment.

LLMs also exhibit promising capabilities in multimodal knowledge extraction. Chen et al. [27] introduced a Chain-of-Thought (CoT)-based distillation strategy for extracting text-image knowledge, achieving state-of-the-art results in multimodal NER and RE. Li et al. [28] developed PGIM, a two-stage multimodal NER framework that leverages ChatChatGPT as an semantic knowledge base and incorporates visual cues from images to enhance entity recognition accuracy.

Existing studies on knowledge extraction in the AEC domain cover rule-based, machine learning, deep learning, and LLM-based approaches. But current knowledge extraction efforts are often designed around specific modalities or isolated extraction tasks, with limited emphasis on unified representation, cross-modal relationship tailored for knowledge graph reasoning and reuse. Future research may therefore focus on more tightly coupled multimodal extraction pipelines, ontology-aware alignment across modalities, and scalable storage frameworks that support both structured querying and downstream reasoning. These directions extend existing approaches toward multimodal knowledge graph construction.

### 2.3 Intelligent Applications of Standard Knowledge

LLMs have been increasingly adopted for knowledge-based applications involving building engineering standards. Zhong et al. [29] fine-tuned a BERT model on a regulatory question-answering dataset, developing an end-to-end QA system with chatbot interaction that enhances querying efficiency for building regulations. Yet research exploring the integration of LLMs with knowledge graphs remains limited. Zheng et al. [22] combined prompt engineering with local knowledge bases to support semantic understanding and text generation in AEC applications. Future research must explore more advanced architectures that tightly couple LLMs with structured knowledge for intelligent standard management.

Standard documents contain extensive knowledge concerning design logic and constraint rules [30]. Although retrieval-based methods enable access to this knowledge, traditional search tools often suffer from inefficiency and redundant results. Knowledge-graph triples-obtained through semantic analysis of clauses-preserve the precise meaning of regulatory requirements, enabling more accurate clause-level retrieval. Retrieval-Augmented Generation (RAG) integrates language models with external knowledge retrieval, supplying LLMs with factual grounding to improve accuracy and reduce hallucinations [31].

Recent studies have explored deeper coupling between knowledge graphs and LLMs. Wang et al. [32] introduced a Chain-of-Knowledge (CoK) mechanism to guide LLMs using structured knowledge and textual evidence. Choudhary et al. [33] proposed a hybrid reasoning framework combining LLMs with knowledge-graph inference via query abstraction, neighborhood retrieval, and query decomposition, achieving breakthroughs in complex logical reasoning. Chao et al. [34] developed GenQA, a Transformer-based QA framework capable of synthesizing information across multiple candidate answers to generate more coherent responses. In 2023, China Communications Construction reported an internal system integrating domain knowledge bases with LLM-RAG pipelines for standard question answering, but without explicit modeling of clause hierarchy or multimodal regulatory constraints [35].

In standard-based applications, accuracy, completeness, and logical rigor are essential. While RAG enhances LLM performance, hallucinations remain a challenge. Knowledge-Augmented Generation (KAG) can be viewed as an extension of RAG, emphasizing dynamic knowledge updates, multi-source integration, and closed-loop optimization. KAG excels in structured knowledge scenarios, producing traceable responses. It has been successfully deployed in specialized domains such as e-government and e-health QA within Ant Group [36]. Wang et al. [37] proposed KnowledgeChatGPT, which enhances LLMs' retrieval and reasoning capabilities by integrating programmatic thinking prompts, diverse knowledge representations, and personalized knowledge bases, effectively addressing entity ambiguity and complex query processing.

With the integration of domain-specific knowledge bases of building engineering standards, KAG holds substantial potential for enabling QA and rule interpretation in engineering practice. Overall, the rapid development of LLMs is unlocking new possibilities for applying building standards in the AEC domain, particularly in intelligent question answering and automated rule-code generation[38].

However, most existing systems still rely on RAG and CoK paradigms, both of which face structural limitations[39]. RAG-based methods mainly perform surface-level semantic retrieval over text fragments, with weak modeling of clause hierarchies and reference structures, resulting in poor handling of formulas, tables, and parameter dependencies. CoK approaches introduce knowledge-guided reasoning, but typically employ simplified or domain-agnostic graphs that fail to formally represent regulatory logic, clause organization, and cross-document relationships, thereby constraining robust and rule-consistent reasoning.

As a result, although early attempts at automated rule-code generation are emerging, systematic and engineering-oriented methodologies for deep LLM-knowledge integration remain limited. Future progress will depend on tighter coupling between LLM and formally structured standard knowledge to support reliable intelligent applications of building standards.

## 3. Methodology

This section proposes the method named BEST-KAG (Knowledge-Augmented Generation for Building Engineering STandards), covering data modeling, automated parsing, and knowledge-augmented generation. The overarching idea is to use a multimodal knowledge graph as the core carrier, uniformly modeling and semantically linking heterogeneous information such as clauses, formulas, tables, and figures scattered across different standard documents. This structured backbone supports subsequent retrieval, reasoning, and downstream applications. We propose an automated knowledge extraction pipeline that combines regular-expression-based parsing with LLM, enabling end-to-end processing from document structure recognition and semantic element labeling to non-text content parsing. This significantly reduces manual intervention and improves both scalability and accuracy in knowledge-base construction. Finally, we design an intelligent question-answering framework BEST-KAG. The overall technical roadmap of the article is shown in Figure 1.

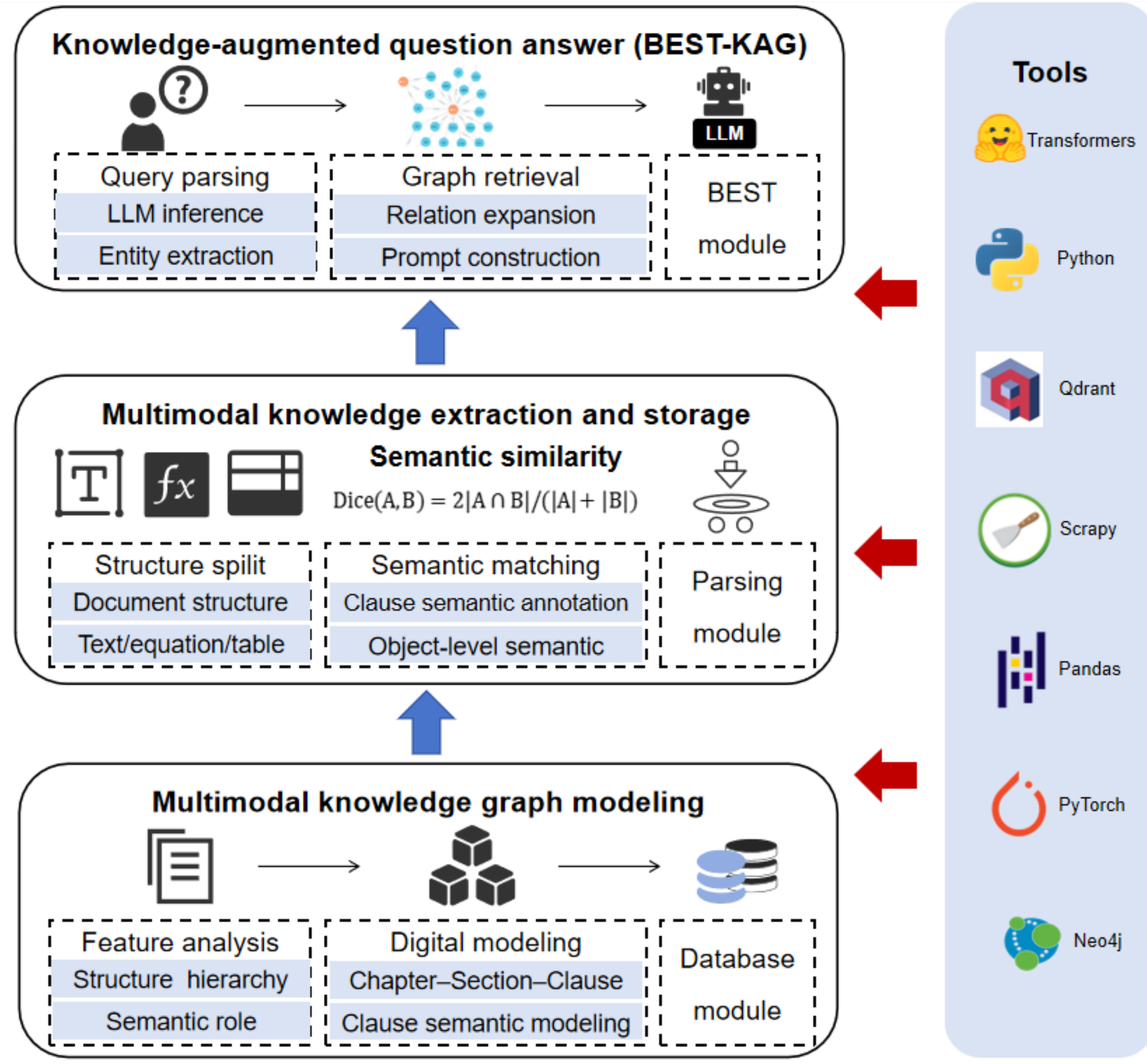

**Fig. 1.** The overall flowchart of the BEST-KAG method and technical route.

### 3.1 KG-Based Multimodal Model for Building Engineering Standards

Within the overall architecture, the first challenge is to represent multimodal standard knowledge in a unified and structured form. This is addressed by the proposed knowledge graph-based multimodal representation model for building engineering standards. Knowledge graphs provide a formal mechanism for organizing heterogeneous semantic information and have been widely adopted for cross-domain knowledge integration and retrieval. Building on this paradigm, we construct a hierarchical and semantically enriched graph model that preserves both the original document structure and the fine-grained regulatory semantics of standard texts.

#### 3.1.1 Hierarchical Modeling of Standard Document Structure

At the document-structure layer, we capture the hierarchical organization of each standard document by explicitly modeling its numbering and organizational system. Each structural level is represented as a node in the graph. A Standard root node corresponds to a single standard document, Chapter nodes represent chapters, Section nodes represent sections, and Clause nodes represent individual clauses. These nodes are connected through hierarchical relations, including hasChapter, hasSection, and hasClause, which preserve the contextual embedding of each clause within the overall document. The overall chapter structure is shown in Figure 2 and Table 3.

To support associative retrieval and regulatory reasoning across different parts of the standards, we further model citation and cross-reference relationships. Specifically, hasReference relations are defined between Clause nodes and other Clause, Section, or Chapter nodes to capture both intra-level and inter-level references. This explicit representation of cross-references enables the graph to encode dependency structures among regulatory provisions and provides a structural foundation for evidence tracing and context-aware query resolution in downstream knowledge-augmented question answering tasks.

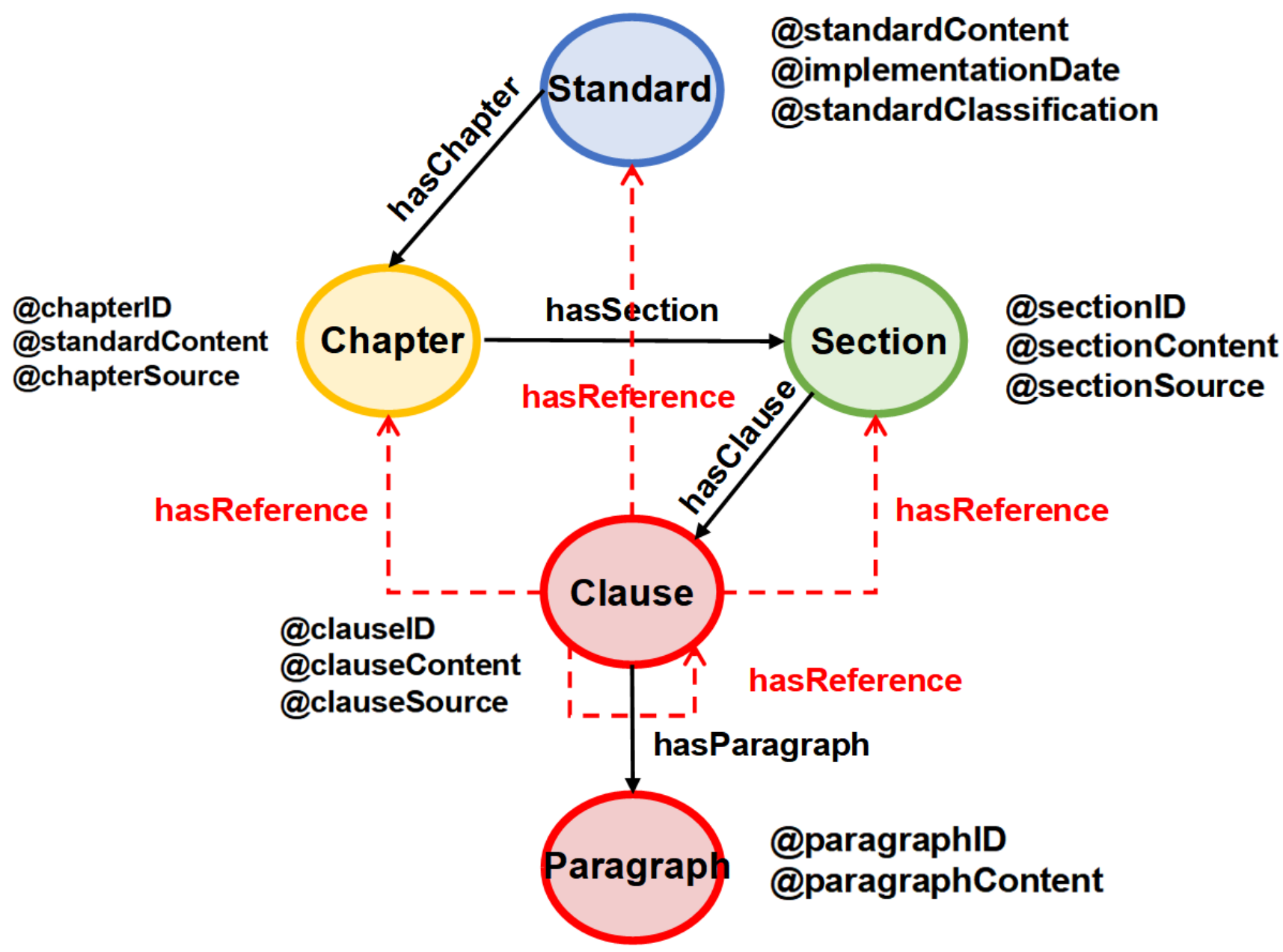

**Fig. 2**. Hierarchical and Citation-Aware Knowledge Graph Schema for Standards.

**Table 1**

Description of node types and attributes in the standard document layer

| Node type | Attribute Name | Property Description |
|---|---|---|
| Standard | @standardID | Standard Sequence Number |
| | @standardName | Standard Name |

| Node type | Attribute Name | Property Description |
|---|---|---|
| | @implementationDate | Publication Year |
| | @standardClassification | Standard Classification |
| Chapter | @chapterID | Chapter ID |
| | @chapterContent | Chapter Title |
| | @chapterSource | Chapter Source |
| Section | @sectionID | Section ID |
| | @sectionContent | Section Title |
| | @sectionSource | Section Source |
| Clause | @clauseID | Clause ID |
| | @clauseContent | Clause Content |
| | @clauseSource | Clause Source |

**Table 2**

Types of relationships in the standard document layer and their explanations

| Relationship type | Starting node | Ending node | Relationship Explanation |
|---|---|---|---|
| hasChapter | Standard | Chapter | Standard Has Chapter |
| hasSection | Chapter | Section | Chapter Has Section |
| hasClause | Section | Clause | Section Has Clause |
| hasParagraph | Clause | Paragraph | Clause Has Paragraph |
| hasReference | Clause | Chapter | Clause References Standard |
| | Clause | Chapter | Clause References Chapter |
| | Clause | Section | Clause References Section |
| | Clause | Clause | Clause Cross-Reference |

### 3.1.2 Multimodal Semantic Modeling at the Clause Level

At the clause-semantics layer, Paragraph nodes serve as the core units. The knowledge graph structure centered around paragraphs is shown in Figure 3. Each Paragraph node is associated with attributes such as @paragraphContent, @paragraphOriginal, and @paragraphSource. To overcome the limitations of prior work that focuses solely on text, tables, figures, and formulas are modeled as subtype nodes Table, Figure, and Equation, respectively. These are tightly linked to their corresponding Paragraph nodes via hasTable, hasFigure, and hasEquation relations. Each multimodal node stores metadata such as @imageID and @equationDefinition to preserve the original structure and support traceability. For tables and figures, we adopt a localization-indexing-description strategy. For tables, we identify headers, row-column structures, and annotations, and use the LLM to extract parameter indicators and inter-variable relationships, yielding structured textual descriptions. For figures, we focus on core information such as geometric relationships and construction details, recognizing graphical symbols, dimension annotations, and textual notes, then converting them into standardized textual explanations with the help of the LLM. Both table and figure descriptions are indexed by unique identifiers and linked to their associated clauses, ensuring semantic alignment between non-text and textual knowledge. At a finer-grained semantic level, Paragraph nodes are connected to Situation nodes via hasSituation, which in turn connect to Object, Property, Value, Function, and Limit nodes via hasObject, hasProperty, hasValue, hasFunction, and toLimit relations. This yields a multi-level semantic network capturing condition-object-property-value-constraint chains. The resulting architecture preserves the original structure of the clauses while enriching them with semantic information that facilitates automated interpretation of standard texts.

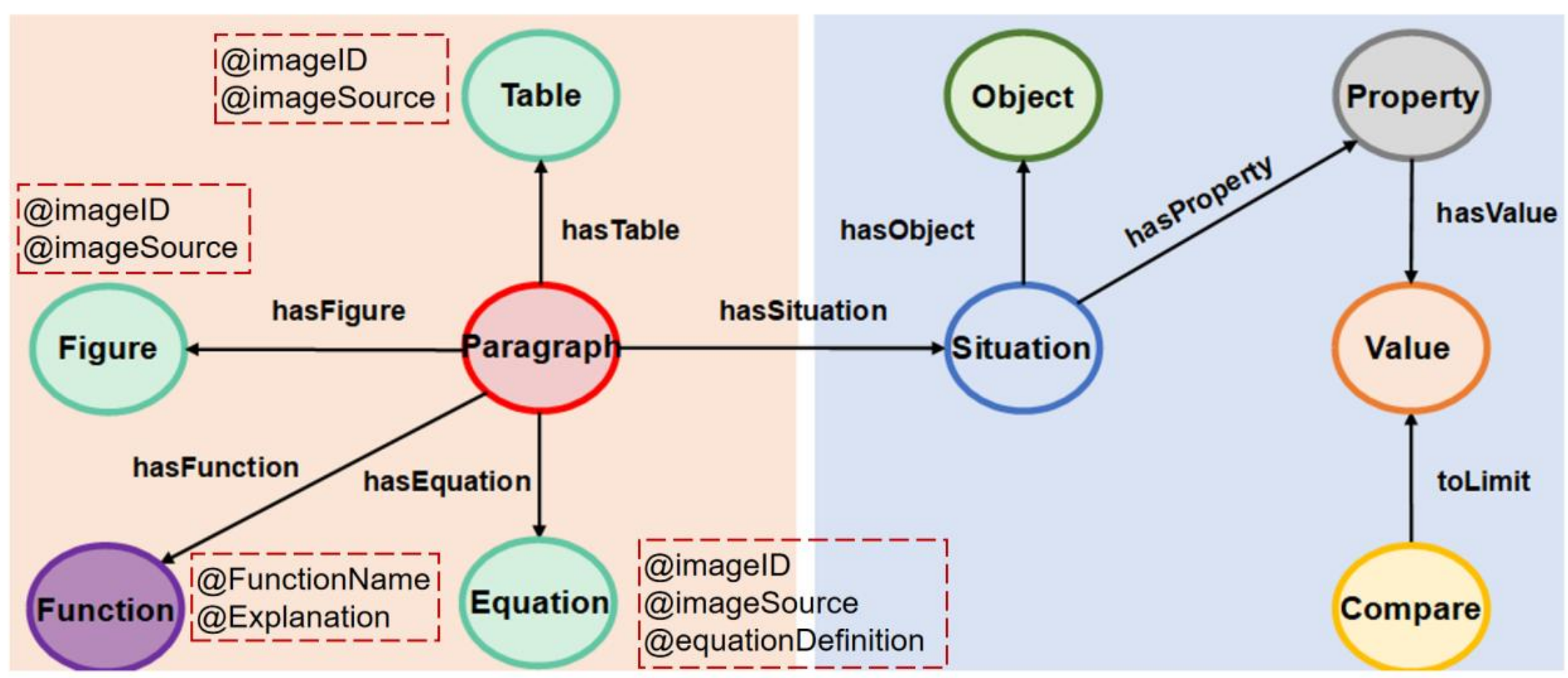

**Fig. 3.** Node types and their interrelationships in the knowledge graph.

### 3.2 Construction of the Multimodal Knowledge Base

Given a clear knowledge organization framework, the next task is to efficiently transform large, complex standard documents into graph data. This is the central goal of the automated knowledge-base construction process. The multimodal knowledge base provides the core infrastructure for digital management and intelligent application of building engineering standards. Its key task is to integrate regular-expression-based preprocessing, LLM-based parsing, and graph-database storage to convert heterogeneous information-text, formulas, tables, and figures-into structured knowledge with well-defined relations. The pipeline encompasses structural decomposition of standard documents, deep parsing of textual and non-textual content, and unified storage of multimodal knowledge, forming an end-to-end technical chain.

#### 3.2.1 Automated Structural and Reference Parsing

Structural decomposition is the foundational step of knowledge-base construction. We employ regular expressions like Figure 4 and Table 4 to automatically extract standard attributes, hierarchical structure, and citation relationships, providing structured inputs for subsequent semantic parsing.

For standard attributes, we design regular expressions based on the normalized naming format of standard documents. A full title typically consists of “standard name + space + standard number,” where the standard number includes a code, a sequence number, and a publication year. By matching these components, we accurately extract attributes such as standard name (@standardContent), standard ID (@standardID), and implementation year (@implementationDate). Using the mapping between standard codes and their types (e.g., GB for mandatory national standards, GB/T for recommended national standards), we further derive the classification attribute (@standardClassification).

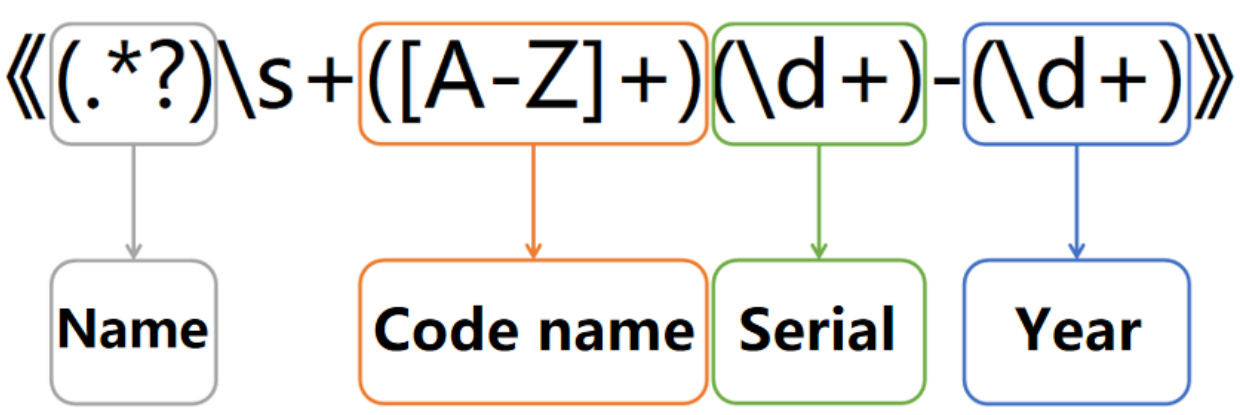

**Fig. 4.** The regular expression used for standard attribute extraction.

For chapter-section-clause information, we design a multi-stage filtering mechanism reflecting the hierarchical numbering scheme. First, we use regular expressions to match lines beginning with numeric chapter identifiers, including common patterns that combine Arabic numerals with Chinese or English separators, while marking non-structural lines with "0." We then construct a three-level title system (chapter, section, clause) by parsing the dot-separated numbering (e.g., "1.2.3" corresponding to Chapter 1, Section 2, Clause 3). Additional refinement rules remove spurious top-level headings with large numeric values, strip page numbers and other pseudo-headings, and correct anomalous paragraphs to improve recognition accuracy. After extraction, we classify the data into chapter-level, section-level, and clause-level CSV files, preserving the complete three-level structure and attaching the standard name to ensure traceability.

**Table 3**

Regular expression that matches the reference format of chapter-section-clause

| Referenced Level | Corresponding Regular Expression |
|---|---|
| Chapter | (code\|standard) Chapter (\d+) |
| Section | (code\|standard) Section (\d+)(\.)(\d+) |
| Clause | (code\|standard)Clause (\d+)(\.)(\d+)(\.)(\d+) |

For citation relationships, we use regular expressions to identify both intra-standard and inter-standard references. Intra-standard references are expressed in forms such as "code, Chapter X," "standard, Section X.X," or "this code, Clause X.X.X." We design specialized patterns to extract chapter IDs (@chapterID), section IDs (@sectionID), and clause IDs (@clauseID) from @paragraphContent, then traverse the hierarchy (chapter → section → clause) in the knowledge graph to locate the referenced nodes and create directed hasReference relations. Inter-standard references are identified by patterns , from which we extract the full title of the cited standard. We then locate the corresponding Standard node in the graph and establish a reference relation, explicitly encoding the dependency structure among standard documents.

### 3.2.2 Semantic Parsing of Textual Content with LLMs

We employ LLMs to semantically decompose and annotate clauses that have undergone structural decomposition, transforming natural language into machine-interpretable semantic labels. This process consists of two progressive stages: clause segmentation and semantic labeling.

We design a structured prompt template for clause segmentation, which consists of five components: task specification, processing steps, quality control rules, handling of ambiguous cases, and few-shot examples. The template explicitly assigns the LLM the role of a professional regulatory text-parsing assistant and constrains it to accurate structural decomposition without semantic reinterpretation. Clause segmentation follows a coarse-to-fine procedure. First, clause-internal enumeration markers are identified as primary segmentation boundaries. Second, the original clause is split into independent subclauses based on these markers. Third, each subclause is further refined using punctuation marks such as periods and semicolons to obtain finer-grained units. The output is normalized into a standardized format in which each segmented unit is placed on a separate line and enclosed in square brackets.

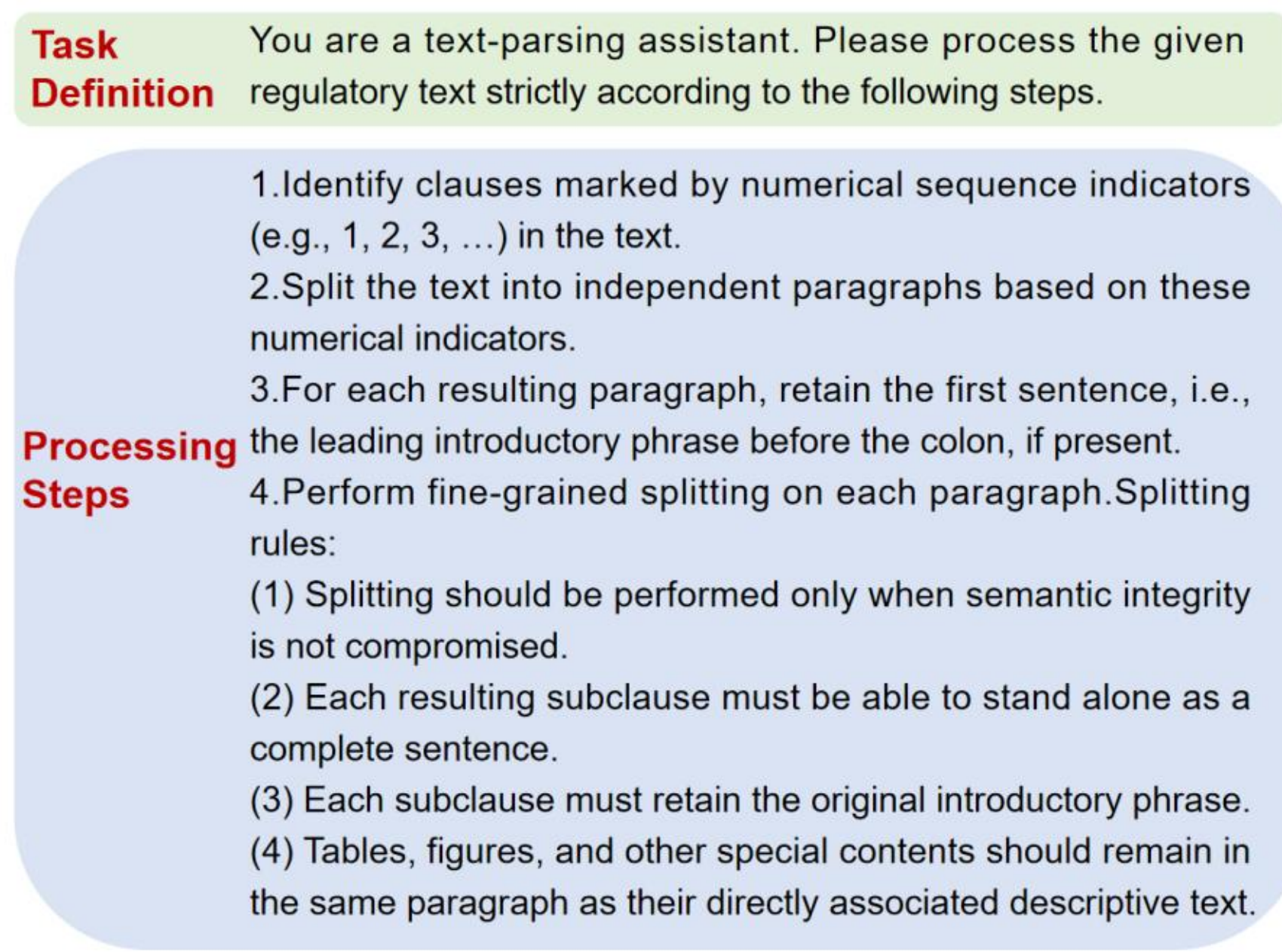

**Fig. 5.** Clause Breakdown Template.

For semantic labeling, we design a prompt template as shown in Fig. 6 based on five core semantic categories in the digital standard model: technical term (obj), comparative relation (cmp), constraint condition (sit), attribute definition (prop), and attribute value (val). The template specifies the extraction task and provides definitions and few-shot examples for each category to clarify their boundaries. It emphasizes that constraint conditions must not be duplicated in other labels and that each element should be assigned to the most appropriate category. The output format requires each category to be listed on a separate line, with items wrapped in square brackets and separated by commas, ensuring coverage of all key semantic information. To enhance accuracy, we adopt a multi-round processing strategy: we validate the initial annotations and, if a subclause contains multiple constraints or technical terms, we further split it into smaller units. For subclauses without explicit constraint conditions, we automatically assign a default condition such as "under all circumstances," ensuring that every subclause is associated with a clearly defined applicability context.

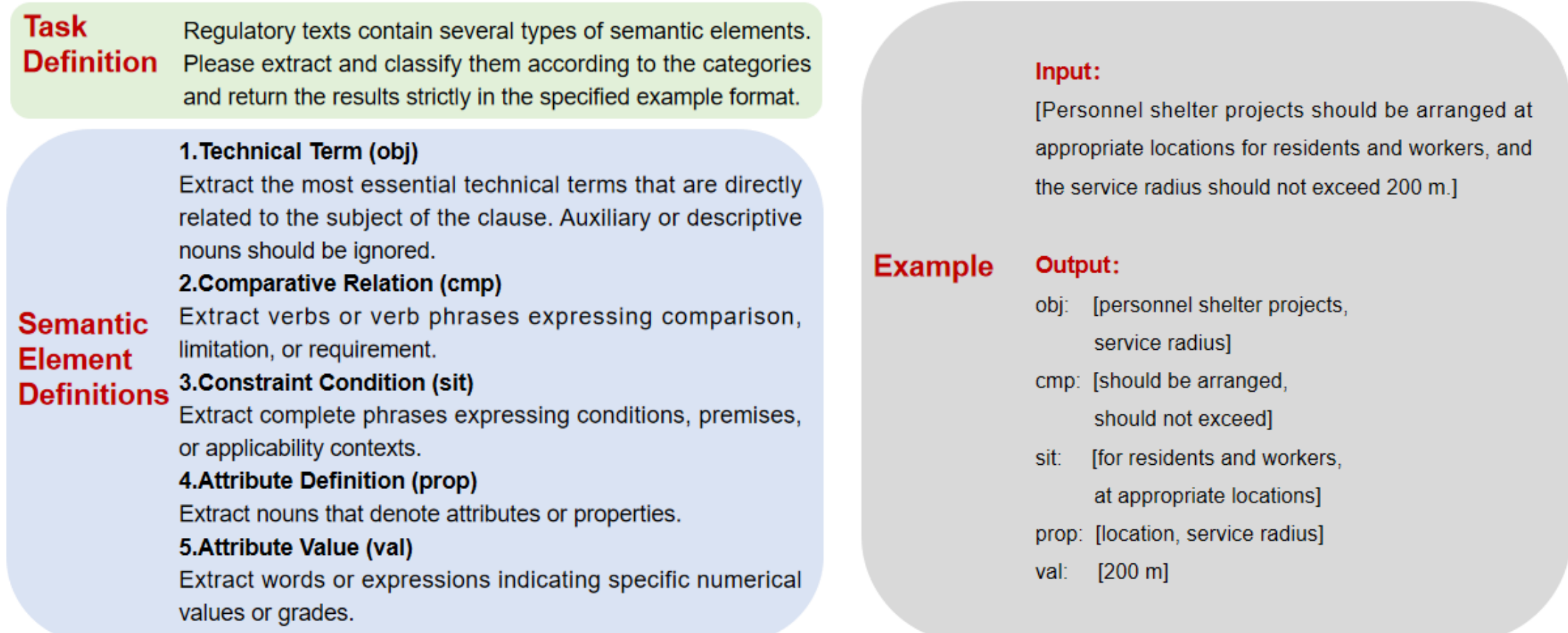

**Fig. 6.** Semantic Annotation Template.

### 3.2.3 Parsing Equation and Table with LLMs

Parsing non-text content (formulas, tables, and figures) requires multimodal processing to achieve machine interpretability. In this work, we employ LLMs to convert original non-text content into structured descriptions.

For formulas, we use multimodal LLMs to translate mathematical expressions from image form into natural language. The core process is as follows. We extract clauses containing formulas from the database, where formulas are stored as Base64-encoded images. These encodings are converted into temporary image files. We then design a prompt template that combines the surrounding textual context with a clear task specification requiring the LLM to produce a professional, human-readable explanation of the formula wrapped in square brackets, containing only the explanatory text. Few-shot examples mapping formulas to their natural language descriptions help the model learn the desired style and level of detail. We invoke the multimodal LLM interface to generate normalized descriptions, which are then written back into the knowledge graph. To ensure quality, we enforce strict output formatting and implement exception handling, including temporary-file management and delay strategies for concurrent access. We select an LLM capable of joint text-image understanding to accurately capture symbols, variable relationships, and computational logic in the formulas.

| | |
|---|---|
| **Task Definition** | Please translate this formula into a natural language interpretation and a LaTeX representation. |
| **Requirem-ents** | 1.The output format must be enclosed in square brackets “[ ]”.<br>2.Only output the natural language interpretation and the LaTeX representation. Do not include any other content. |
| **Few-shot Examples** | Formula:<br>$S \le R/\gamma RE$<br>The action effect shall not exceed the load-bearing capacity of the structural component divided by the seismic adjustment factor of resistance. [S \le \frac{R}{\gamma_{RE}}] |

**Fig.7.** Formula content interpretation template.

### 3.2.4 Relationship Extension through Clause-Level Semantic Similarity

We adopt Neo4j as the primary graph database platform and use Py2neo to integrate textual and non-textual content within a unified entity-relation model that accurately reflects the structure of standard knowledge.

For textual content, we use a layered import strategy. Based on the chapter-section-clause CSV files generated in the structural decomposition stage, we import Standard, Chapter, Section, and Clause nodes into Neo4j via Py2neo scripts, constructing the document-layer graph with hierarchical relations (hasChapter, hasSection, hasClause). Using the semantic annotation results, we then create Paragraph, Situation, Object, and other semantic-layer nodes via Cypher queries, linking them through relations such as hasParagraph and hasSituation. Each node stores its textual content in a Content attribute, preserving completeness and semantic similarity. To support semantic similarity computation, we leverage the APOC plugin in Neo4j to compute the Dice similarity coefficient between texts. The Dice similarity for two texts A and B is defined as

$$Dice(A,B)=2|A\cap B|/(|A|+|B|)$$

where $|A \cap B|$ denotes the number of shared characters between the two texts, and $|A|$ and $|B|$ denote the total number of characters in each text. The coefficient ranges from 0 to 1, with higher values indicating greater similarity.

We computing similarity between the Content attributes of Object nodes connected via hasSituation and hasObject. When the weighted semantic similarity between Object nodes exceeds a threshold of 0.8, we create a similarTo relation between them using the Merge command. This allows us to accurately identify clauses with high semantic similarity and visualize their relationships in the graph. As shown in Fig.8, "Grade HRB400 steel" and "HRB400 steel" (where the "Grade" qualifier does not alter the core identity of the steel itself) are accurately identified and linked.

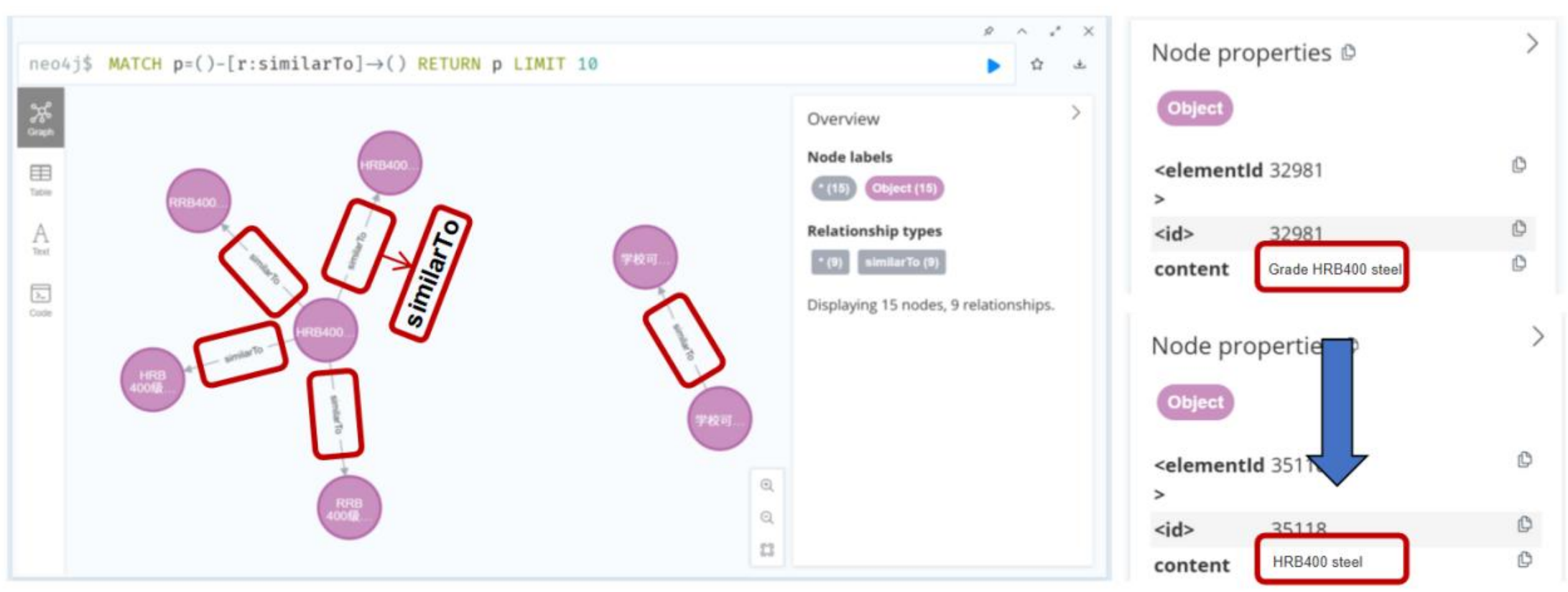

**Fig. 8.** An example of matching the relationship between nodes based on semantic similarity.

For non-text content, storage proceeds in three stages: initialization, relation mapping, and node classification. In the initialization stage, we use Py2neo to batch-import non-text entries as generic Image nodes with attributes imageID (identifier) and imageSource (source standard). In the relation-mapping stage, we match paragraphSource with imageSource and paragraph_imageSource with imageID to link Image nodes to their corresponding Paragraph nodes. In the classification stage, we reclassify Image nodes into Table, Equation, or Figure nodes based on patterns in paragraphContent, and establish hasTable, hasEquation, and hasFigure relations accordingly. For Equation nodes, we store the LLM-generated natural language description in the equationDefinition attribute. For Table and Figure nodes, we link them to their structured descriptions via attributes, thereby achieving semantic storage of non-text content. In this way, we construct a unified knowledge graph that integrates both textual and non-textual information, providing a robust data foundation for knowledge linkage and applications. On top of this multimodal knowledge base, we can further build user-oriented intelligent services, most notably the KAG-based question-answering framework that directly supports engineering practice.

### 3.3 KG-Augmented Question Answering

Building on the multimodal knowledge base described above, we develop a BEST-KAG framework to enable intelligent question answering for building engineering standards. The core idea of the framework is to combine the precise information retrieval capabilities of the knowledge graph with the natural language generation capabilities of LLM, thereby producing accurate, context-rich, and traceable responses.

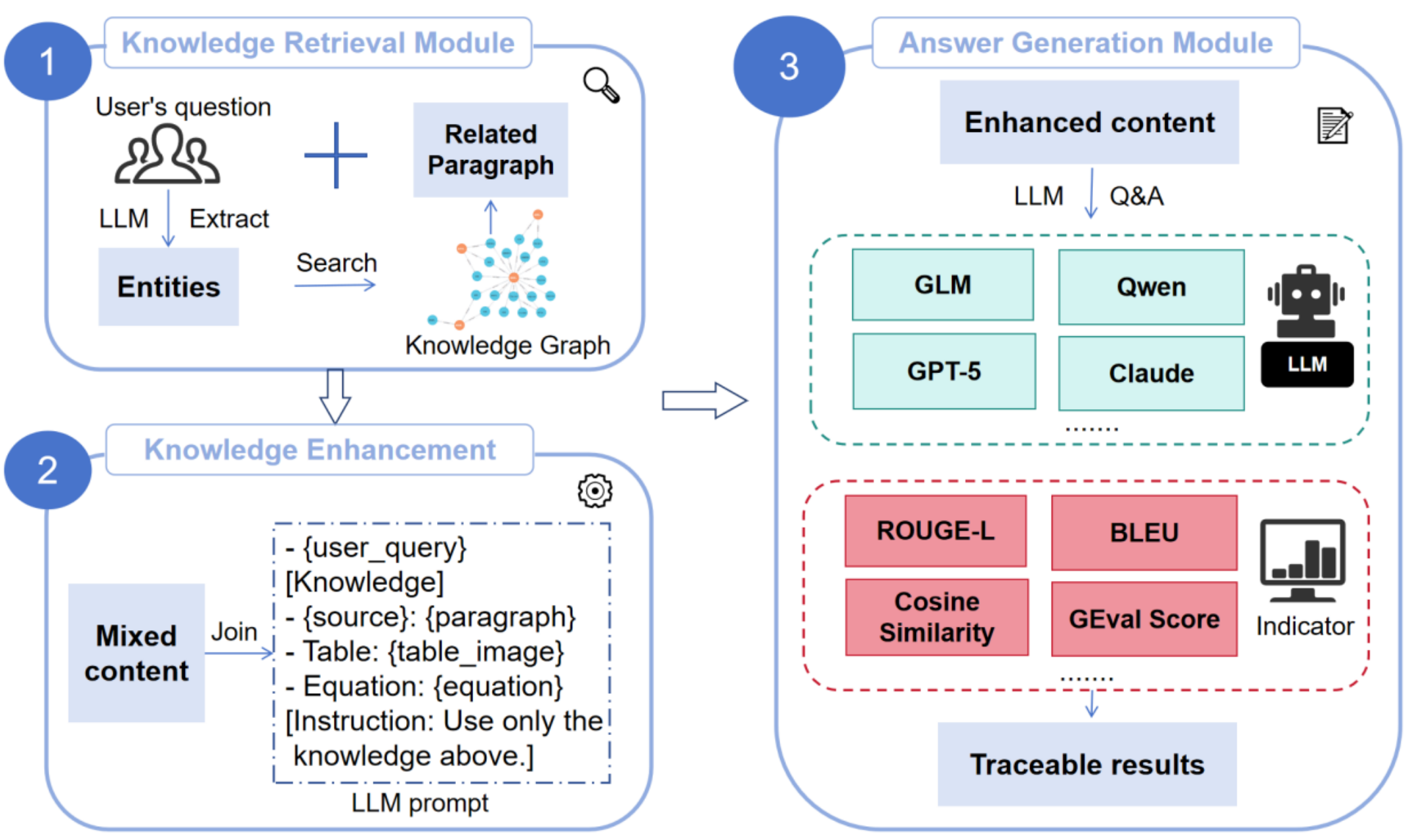

**Fig. 9.** BEST-KAG Knowledge-Augmented QA Workflow.

The overall architecture consists of four components: knowledge retrieval, knowledge augmentation, generation model, and answer presentation with feedback.

In the knowledge retrieval module, the system executes precise graph-based queries according to the user's input. Specifically, an LLM is first used to extract one or more key entities from the query. For each extracted entity, the system locates the corresponding Object nodes in the graph, then traces upward through hasObject relations to the associated Situation nodes, and further through hasSituation relations to the Paragraph nodes. For each Paragraph node, attributes such as paragraphContent, paragraphOriginal, and paragraphSource are retrieved using Cypher queries. In addition, the system retrieves Paragraph nodes linked through similarTo relations, as well as Equation, Figure, and Table nodes linked through hasEquation, hasFigure, and hasTable relations. These supplementary nodes provide semantically similar clauses, formula definitions, graphical information, and tabular data. Base64-encoded images are decoded into JPG format to obtain complete multimodal context. Through this process, the system compiles a comprehensive set of contextual knowledge relevant to the user's query.

```
MATCH (o:Object)
WHERE any(entity IN $entities WHERE
    o.content = entity OR
    o.content CONTAINS entity
)
WITH COLLECT(DISTINCT o) AS originalObjects
```

1.Fetch original Objects matching entities

```
MATCH (o:Object)-[:similarTo]-(similarObj:Object)
WHERE o IN originalObjects
WITH originalObjects + COLLECT(DISTINCT similarObj) AS allObjects
```

2.Expand Objects via similarTo relations

```
UNWIND allObjects AS obj
MATCH (obj)<-[:hasObject]-(:Situation)<-[:hasSituation]-(p:Paragraph)
"""
```

3.Retrieve related Paragraphs

**Fig. 10.** A code example for database query search.

In the knowledge augmentation module, the retrieved information is organized and integrated into a structured input for the LLM. Textual knowledge is concatenated into a coherent context string, while associated images are uploaded as multimodal inputs. This step ensures that the LLM receives a unified and semantically enriched knowledge context rather than isolated fragments.

```
blocks = [
{"text": "Please answer based solely on the standard provisions and
pictures provided. It would be better if you could provide the source
(standard name / chapter number)."} ]

blocks.append({"text": "【检索知识】\n" + knowledge_text})

blocks.append({"text": f"【用户问题】\n{query}"})

blocks.append({"image": as_data_uri(t)})
```

**Fig. 11.** Enhanced prompt after acquiring knowledge.

In the generation module, the system invokes the LLM API again to synthesize an expert-level response. The integrated knowledge context is combined with the user's original query and passed to the model, guided by domain-specific prompt templates designed to ensure accuracy, technical rigor, and readability. These prompts instruct the LLM to generate responses that are not only factually grounded in the retrieved clauses but also presented in a professional style suitable for engineering practice.

## 4. Experiments

### 4.1 Knowledge Base Construction

Building on the digital modeling and automated parsing methods proposed in the previous chapters, this study constructed a multimodal standard knowledge base covering 251 building engineering standards, spanning structural design, fire safety, building materials, construction methods, and other major subdomains. The instantiated Neo4j knowledge graph contains 171,652 nodes and 310,914 edges. Among all node types, Paragraph nodes account for the largest proportion, reflecting their role as carriers of the core textual content. They are followed by semantic-layer nodes such as Object and Situation, which encode refined semantic elements extracted from the clauses.

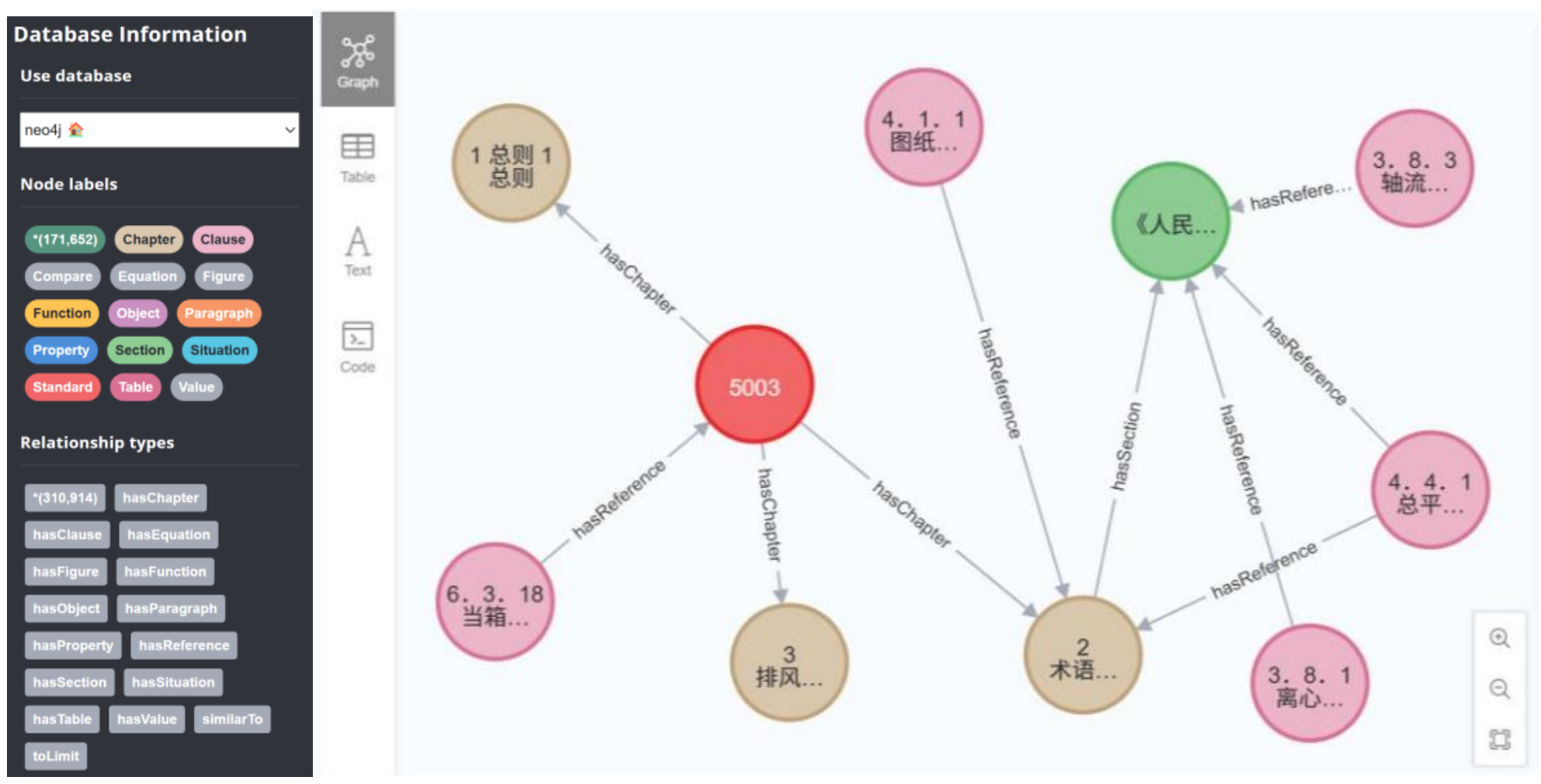

**Fig. 12**. Subgraph visualization of the building engineering standards knowledge graph.

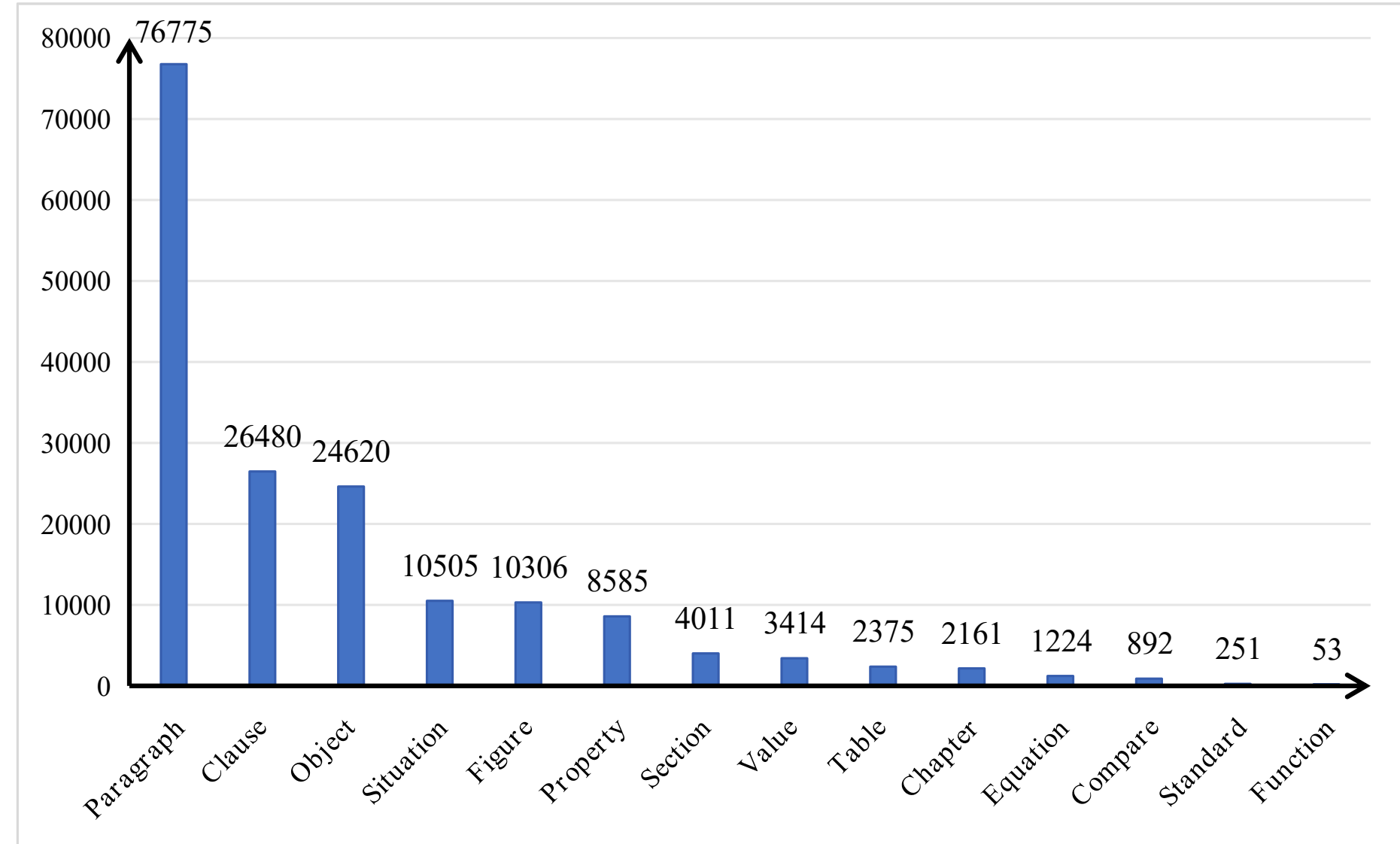

**Fig. 13**. Node type distribution in the knowledge graph.

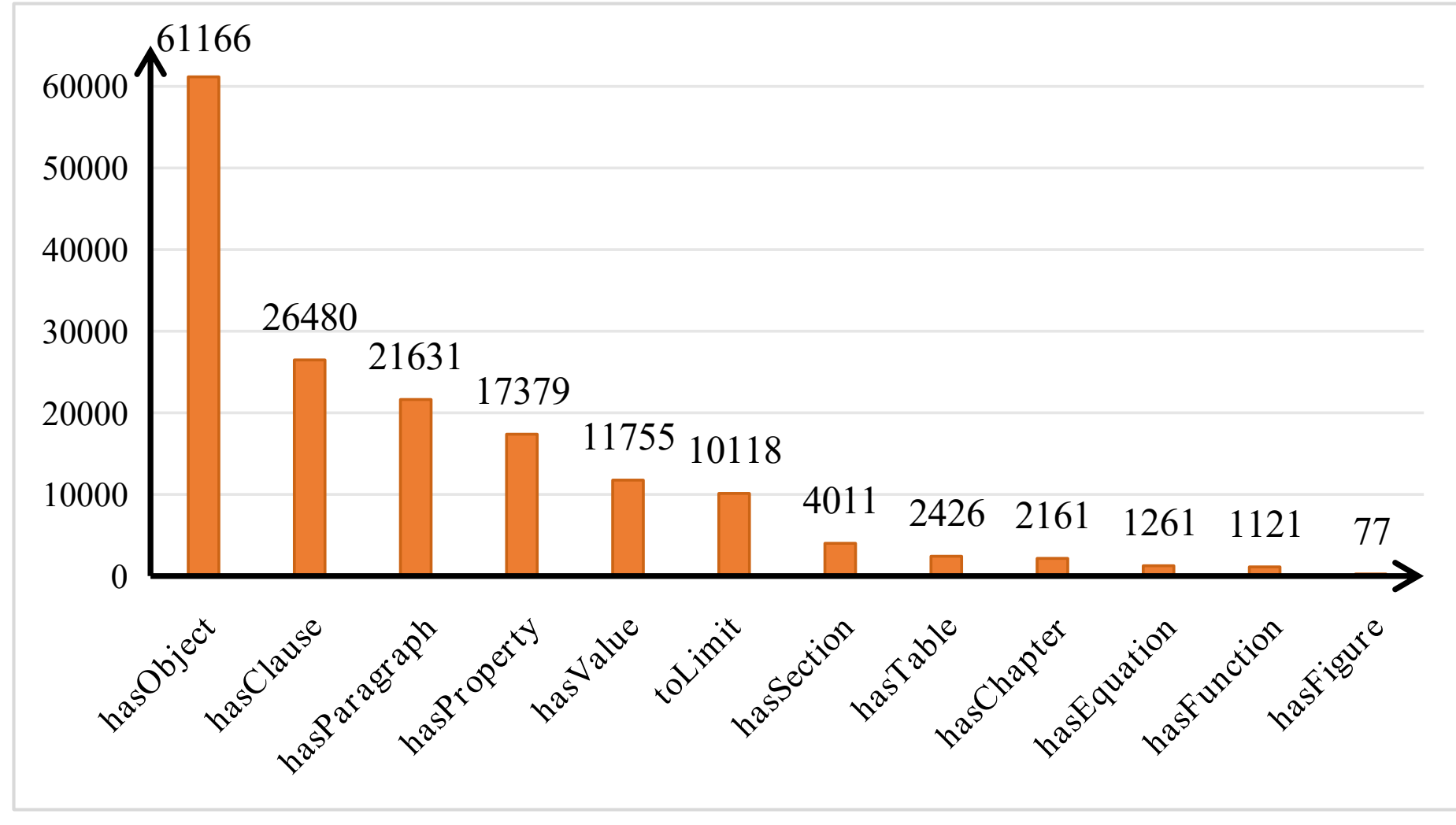

**Fig. 14.** Relationship type distribution in the knowledge graph.

In terms of multimodal coverage, the knowledge base contains 2,375 table nodes, 1,224 equation nodes, and 53 function nodes. All non-textual elements are explicitly linked to their corresponding Paragraph nodes and include either structured representations or natural-language explanations generated through the proposed LLM-based parsing pipeline.

Manual sampling indicates that the hybrid extraction method combining regular-expression parsing and LLM-based semantic analysis achieves over 90% precision in the “Object” and “Situation” categories. The few remaining errors mostly relate to unit conversion or implicit condition interpretation and can be corrected via prompt refinement or post-processing heuristics. Relative to traditional manual annotation pipelines, the proposed approach significantly reduces human labor costs and supports incremental updates, allowing the knowledge base to be rapidly synchronized as new standards are released or existing ones are revised.

From a performance standpoint, the structured storage schema and optimized indexing ensure that most Neo4j queries-including full-text search and multi-hop traversal-return results in under one second, fully satisfying the real-time requirements of the subsequent question-answering framework. This high responsiveness provides a solid foundation for knowledge-driven applications built on top of the multimodal database.

**4.2 Experiment Design**

To ensure fair and comprehensive evaluation, we include four mainstream LLM families representing both text-only and multimodal variants: Qwen-max (text) / Qwen-vl (multimodal), ChatGLM-4-plus (text) / ChatGLM-4v (multimodal), ChatGPT-5, and Claude 3.5. All models are executed under identical conditions (Windows 11, Python 3.10, Neo4j 5.8), using precisely the same prompt templates, sampling parameters, and length constraints to ensure comparability.

We compare two question-answering methods:

1)Text : The model generates answers directly from the prompt.

2)BEST-KAG(our): The model generates answers using BEST-KAG system.

All methods operate within the same environment and pipeline. Batch evaluation scripts handle Excel input/output, error recovery, and multi-model scheduling; graph-side index structures ensure efficient retrieval of Object nodes and traceable evidence. Strict evidence-size limits and controlled Cypher queries guarantee reproducibility, comparability, and statistical robustness.

**4.3 Performance Evaluation**

To assess the effectiveness of the proposed BEST-KAG framework in real building-standard scenarios, we compile a dataset of 140 question-answer pairs, distributed across seven categories (20 per category) to avoid category bias and ensure fair comparison across models.The categories are designed to cover the main types of knowledge and reasoning involved in engineering standards: definition questions target conceptual understanding of technical terms; numerical requirement questions focus on extracting and verifying quantitative constraints; methodological rule questions assess procedural and conditional reasoning; material specification questions examine fine-grained prescriptive details; application scenario questions simulate real-world contextual interpretation across multiple clauses; true-false judgment questions evaluate logical verification and cross-clause consistency; Table/Equation represents the problems related to multimodal data as defined in the standard. Each record contains a question and a standard answer to support batch-aligned evaluation.

**Table 4**
Design examples of seven types of questions and standard answers

| Question Type | Question | Standard Answer |
|---|---|---|
| Definition | What is the definition of *sag* in the *Code for Design of 66 kV and Below Overhead Power Transmission Lines (GB 50061-2010)*? | Sag refers to the maximum vertical distance between the conductor and the straight line connecting the conductor suspension points within one span of an overhead line. |
| Numerical Requirement | What is the required service radius for personnel shelter projects? | Personnel shelter projects should be arranged at appropriate locations for residents and workers, and the service radius should not exceed 200 m. |
| Method / Regulation | What requirements should the capacity of the battery bank satisfy? | Clause 3.7.4 specifies that the capacity of the battery bank shall meet the following requirements: (1) For manned substations, the battery capacity shall support a full-station outage for 1 hour... |
| Material specification | How should raw materials for shotcrete be selected? | Clause 3.4.1 specifies that: (1) Ordinary Portland cement shall be used, with a strength grade not lower than 32.5; damp, expired, or caked cement is strictly prohibited. (2) Hard, clean medium sand ... |
| Application Scenario | What requirements should be met for the installation height and orientation of loudspeakers? | Clause 4.3.2 specifies that the installation height and horizontal and vertical orientation of broadcast loudspeakers shall be determined based on acoustic design and site conditions... |
| True / False Judgment | Is it correct that lap joints of self-adhesive polymer-modified bitumen waterproof membranes should be heated with cold air during low-temperature construction? | False. Clause 6.4.5 specifies that during low-temperature construction, lap joints of self-adhesive polymer-modified bitumen waterproof membranes should be heated with hot air. |

| Table/Equation | When the measured non-heating energy consumption index of a public building equipped with a thermal energy storage (ice storage) system is $e_0$, how should the corrected value $e'$ be calculated? On what basis is the correction coefficient σ determined? | The corrected value $e'$ should be calculated using the formula $e' = e_0 \times (1 - \sigma)$, where $e_0$ is the measured non-heating energy consumption index of the public building equipped with a thermal energy storage (ice storage) system [kW·h/(m²·a)]. The correction coefficient σ is determined in accordance with Table 5.3.5, based on the proportion of the building's actual annual thermal storage (ice storage) cooling capacity to the building's total annual cooling demand. When this proportion is less than or equal to 30%, σ shall take the corresponding value specified in the table. |
|---|---|---|

All methods share identical prompts and output formats. The evaluation employs five objective metrics:

·**BLEU-1 / BLEU-2**[40]: n-gram overlap between generated and reference answers

·**ROUGE-L**: longest common subsequence match

·**Cosine Similarity**: semantic embedding similarity

·**Expert Score**: given by ten domain experts based on a unified rating standard: numerical accuracy, content completeness, logical coherence, and traceability. The maximum score is 1.0, and each indicator is evenly distributed. The final score is calculated by averaging the scores of each indicator and the expert ratings.

For each model and question category, we compute averaged scores across all metrics. Expert serves as the primary indicator of answer quality, while the other metrics serve as supporting evidence of semantic fidelity and content overlap.

## 5. Results and Analysis

### 5.1 Model performance

Under consistent inference settings and using the proposed 140-question benchmark, we systematically compare four mainstream LLM families in the text-only method versus the KAG-enhanced method. Metrics include BLEU-1, BLEU-2, ROUGE-L, cosine similarity, and Expert, with Expert as the main performance indicator.

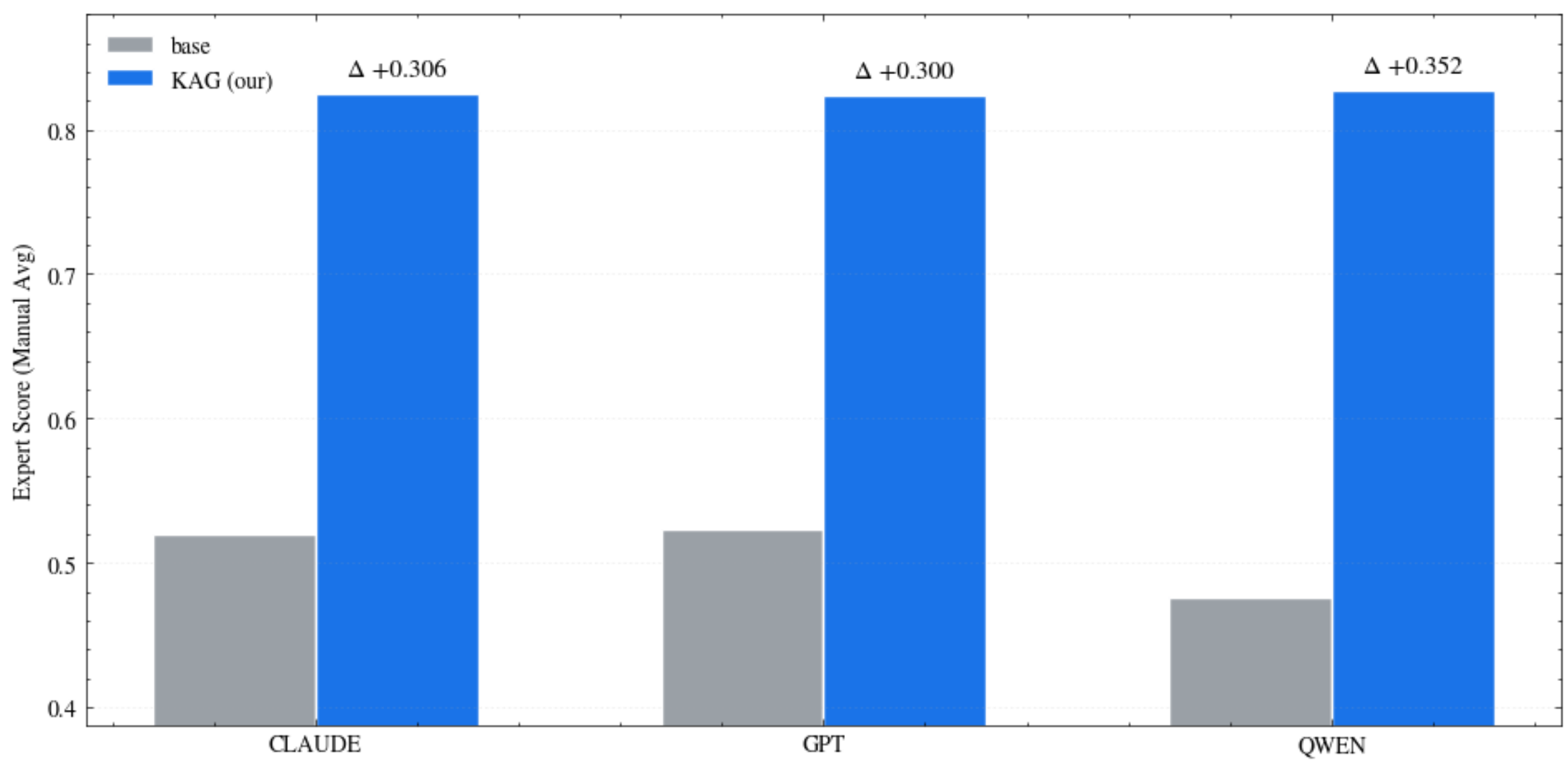

**Fig. 15.** The Expert comparison of the base of each model and the BEST-KAG system.

As shown in Fig. 15, BEST-KAG leads to substantial improvements for all compatible models. The Expert score of Claude increases from 0.52 to 0.83, while ChatGPT improves from 0.52 to 0.82. Qwen shows the largest gain, rising from 0.48 to 0.83. In relative terms, the performance improvements reach 58.84%, 57.21%, and 74.01%, respectively, demonstrating that the proposed framework consistently enhances model performance across different LLM backbones. In contrast, the ChatGLM family exhibits severe degradation in the BEST-KAG method: its Expert score drops from 0.3104 to 0.0055, indicating near-total failure. This suggests that ChatGLM has poor compatibility with the current BEST-KAG pipeline, likely due to sensitivity to long-context noise.

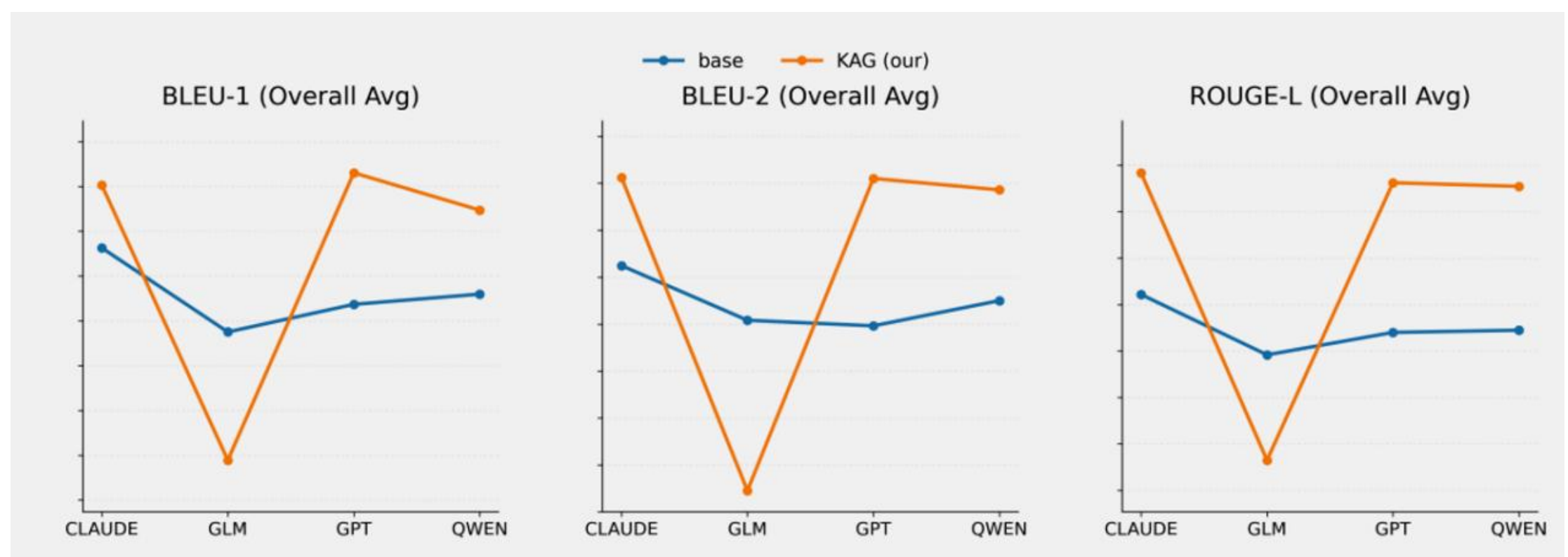

**Fig. 16.** The comparison of the base of each model and the BEST-KAG system.

Across BLEU-1, BLEU-2, and ROUGE-L, the differences between BEST-KAG and the baseline are modest and sometimes irregular. This is expected, since these metrics are fixed-form n-gram overlap measures that primarily reward surface-level similarity to the reference and can undervalue semantically correct, clause-compliant paraphrases. In standards QA, correctness hinges on whether the answer is supported by the cited clauses rather than on reproducing an identical wording; consequently, BLEU/ROUGE are better interpreted as indicators of lexical alignment, not as definitive evidence of accuracy. We therefore place greater weight on expert evaluation and traceability, which directly assess compliance with authoritative standards and the validity of the supporting evidence.

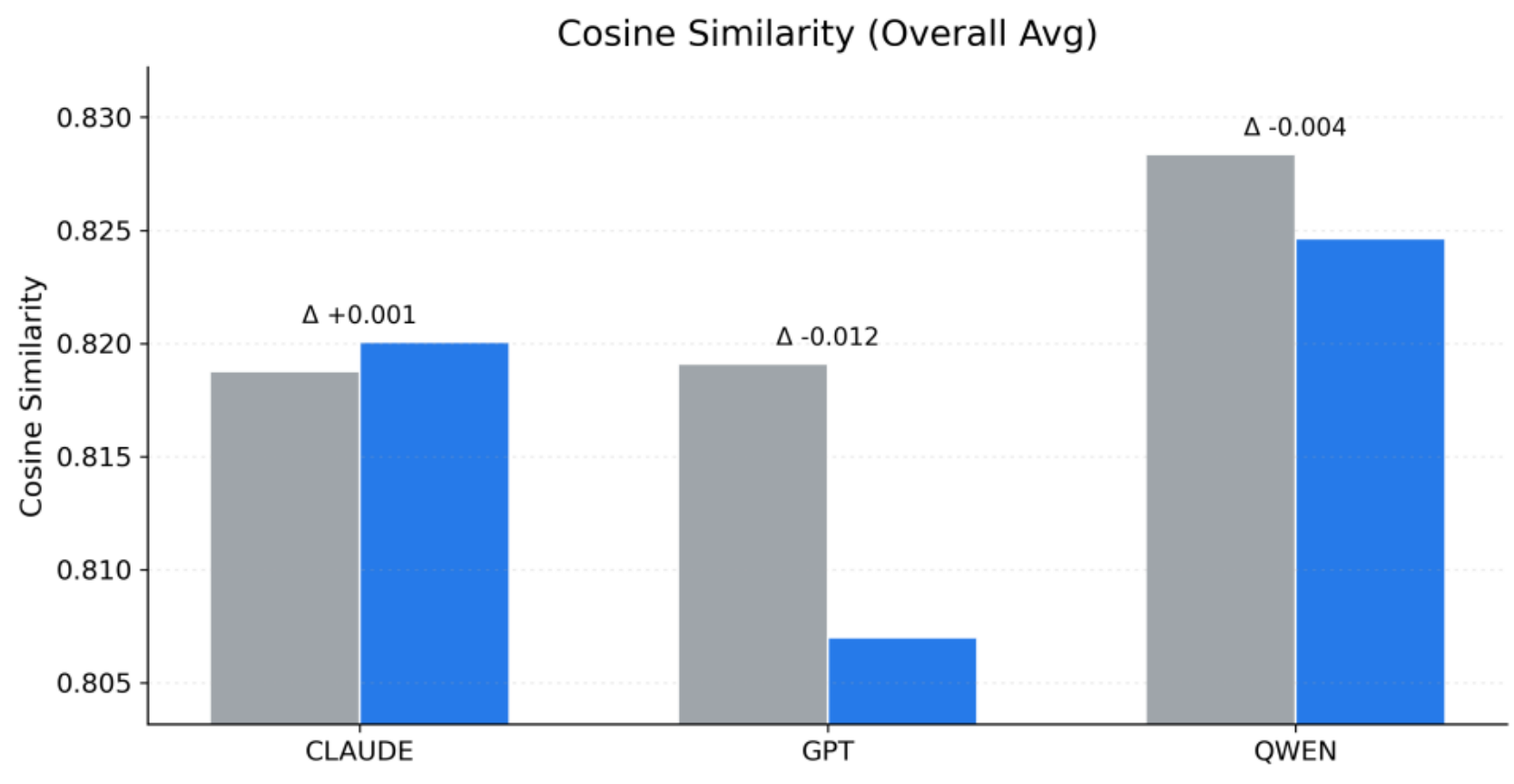

**Fig. 17.** The Cosine Similarity comparison of the base of each model and the BEST-KAG system.

Semantic-level cosine similarity changes are smaller. All four models already achieve high base-method semantic similarity (~0.8). With KAG, Claude, ChatGPT, and Qwen show curves nearly identical to the baseline or only slightly lower, while ChatGLM experiences a sharp drop. This pattern reflects that BEST-KAG primarily improves grounding, accuracy, and textual fidelity, rather than shifting high-level semantics. Though cosine similarity improves only marginally, Expert confirms substantial gains in professionalism, completeness, and logical structure. For ChatGLM, however, the introduction of extensive evidence appears to introduce noise that disrupts decoding, leading to answers that diverge substantially from the reference.

### 5.2 Performance Comparison Across Different Question Types

Based on the category-wise Expert scores, BEST-KAG changes model behavior most in question types where correctness is determined by strict constraint execution rather than fluent free generation. The evidence is clearest in number, method, material, and Table or Equation questions, which require the model to locate the governing clause, identify applicability conditions, and then apply an explicit limit, step, or parameter mapping. After grounding, the models shift from producing plausible but incomplete answers to producing clause-aligned answers that preserve boundary conditions and technical qualifiers. This mechanism is reflected by large score increases in these categories across multiple models, such as ChatGPT in method from 0.36 to 0.81 and in number from 0.43 to 0.82, Qwen in number from 0.42 to 0.86, and Claude in material from 0.44 to 0.89.

Table or Equation questions illustrate why grounding matters beyond surface-level fluency.

Without targeted evidence, models can easily conflate similar tables, misread which variable a coefficient belongs to, or apply a formula under the wrong scope statement. With BEST-KAG, the answer is anchored to the specific table or formula referenced by the clause, so the model is less likely to select an adjacent but incorrect item and more likely to reproduce the correct parameter correspondence. This is consistent with the large improvements observed for Claude from 0.45 to 0.78 in Table or Equation, indicating that the remaining difficulty is often not mathematical manipulation but evidence linkage and correct selection.

In contrast, categories that can be answered with shorter self-contained statements exhibit more model-dependent headroom. True-false shows a ceiling effect for Claude, which already performs strongly at 0.77 and rises modestly to 0.83, while ChatGPT and Qwen improve substantially from 0.77 to 0.98 and from 0.53 to 0.93, respectively. This divergence suggests that for models with weaker baseline calibration, retrieved clauses help suppress overconfident guessing and force a decision supported by an explicit normative statement, whereas for models that already handle such items well, additional evidence yields smaller gains. Definition follows a similar pattern, with the strongest change for Qwen from 0.53 to 0.87, implying that grounding is particularly helpful when definitions in standards are formal, conditional, and easily distorted by paraphrasing. Overall, the category-wise results support a causal interpretation that BEST-KAG primarily improves accuracy by tightening the answer space to what the standard explicitly permits, with the largest benefits appearing when the task depends on precise constraints, procedural completeness, and correct table or formula grounding.

In sharp contrast, the ChatGLM multimodal model experiences near-total performance collapse in all categories: six of the seven question types record Expert scores below 0.01 under the BEST-KAG method, indicating an inability to handle long-context evidence or multi-hop retrieved inputs.

**Table 5**

The performance of different models in different types of problems

| **model** | **system** | **Table/Equation** | **true-false** | **definition** | **application** | **number** | **method** | **material** |
|---|---|---|---|---|---|---|---|---|
| Claude | base | 0.45 | 0.77 | 0.58 | 0.51 | 0.47 | 0.43 | 0.44 |
| | KAG (our) | 0.78 **(+0.33)** | 0.83 **(+0.06)** | 0.83 **(+0.25)** | 0.82 **(+0.31)** | 0.76 **(+0.29)** | 0.83 **(+0.40)** | 0.89 **(+0.45)** |
| ChatGPT | base | 0.50 | 0.77 | 0.63 | 0.47 | 0.43 | 0.36 | 0.52 |
| | KAG (our) | 0.57 **(+0.07)** | 0.98 **(+0.21)** | 0.88 **(+0.25)** | 0.84 **(+0.37)** | 0.82 **(+0.39)** | 0.81 **(+0.45)** | 0.88 **(+0.36)** |
| Qwen | base | 0.49 | 0.53 | 0.53 | 0.52 | 0.42 | 0.44 | 0.42 |
| | KAG (our) | 0.78 **(+0.29)** | 0.93 **(+0.40)** | 0.87 **(+0.34)** | 0.78 **(+0.26)** | 0.86 **(+0.44)** | 0.77 **(+0.33)** | 0.83 **(+0.41)** |

**Table 6**

Percentage improvement (%) from base to KAG across different models and semantic categories

| Category | ChatGPT | Claude | Qwen |
|---|---|---|---|
| Table/Equation | 14.00% | 73.33% | 59.18% |
| application | 78.72% | 60.78% | 50.00% |
| definition | 39.68% | 43.10% | 64.15% |
| material | 69.23% | 102.27% | 97.62% |
| method | 125.00% | 93.02% | 75.00% |
| number | 90.70% | 61.70% | 104.76% |
| true-false | 27.27% | 7.79% | 75.47% |

Taken together, the results indicate that for ChatGPT and Qwen-the strongest and most stable families-introducing multimodal knowledge-graph evidence systematically improves overall QA quality for engineering standards. The improvements are most significant for tasks that heavily rely on clause-level details such as numerical calculation, procedural requirements, and material/construction specifications. This trend is reflected consistently across BLEU-1, BLEU-2, ROUGE-L, and Expert metrics.

However, the effect of BEST-KAG is task-dependent: for definition-level and certain scenario-based questions, simply adding more evidence does not necessarily improve performance, suggesting that these tasks benefit less from explicit clause retrieval.From an engineering-standards perspective, category differences mainly come from how the answer is formed. Definition and some system questions often have short answers that can be stated directly from a single clause, so additional context may not always be necessary. In contrast, numerical, method, and material questions typically require collecting multiple constraints (e.g., limits, conditions, exceptions) and presenting them in an organized way; therefore, they benefit more from clause retrieval and cross-reference expansion. Table/Equation questions are a special case: they rely on correct alignment between clause text and multimodal evidence, so retrieval precision and evidence ranking are more important than simply increasing the amount of context.

**5.3 Analysis of typical text-based cases**

This section selects typical problem cases for analysis. First, we consider a representative text-based question: What regulations should be followed for the selection and arrangement of indoor supply and exhaust vents? This task has strong requirements for structure and standardization: the correct answer must be presented in a standardized list format, retain clear numerical thresholds, and must be traceable to specific clauses, rather than providing general design guidance.

Across all three model families, the base setting typically yields descriptive but incomplete responses. These answers reflect practical HVAC knowledge but often omit clause numbering, collapse multiple conditions into a single narrative, or fail to report quantitative constraints, which limits their expert evaluation in method- and application-oriented categories.

**Reference answer**

…The outlet air velocity should not exceed 3 m/s. The exhaust outlet suction velocity should not exceed 3 m/s. Supply and exhaust outlets should not be arranged facing each other, and the distance should be no less than 1.0 m. …

**base**

…Select suitable outlet type and size based on room function, area, height, and airflow organization requirements. Avoid airflow short-circuiting. The outlet should have an air-volume adjustment function...

**KAG（our)**

…The outlet air velocity should not exceed 3 m/s. The exhaust outlet suction velocity should not exceed 3 m/s. Supply and exhaust outlets should not be arranged facing each other, and the distance should be no less than 1.0 m…

**Fig. 18**. Comparison of text-case the results of the base system with KAG system.

With KAG enabled, the system operates at the level of regulatory units rather than free text. The entity extraction module identifies domain-specific terms such as Indoor supply air outlet, which are then used to retrieve the corresponding clause from the structured standard corpus. The generation stage reconstructs the response by aligning each sentence with a retrieved regulatory item, allowing the full five-point structure of Clause 4.4.4 to be reproduced together with explicit velocity and spacing limits and a formal citation. This mechanism explains the consistent score improvements observed for Claude and Qwen in method, application, and material-related categories, where completeness and procedural fidelity are critical.

At the system level, the results also revealed a dimension of robustness. In the GPT case, the extracted entity airflow organization received a higher retrieval weight than the target entity, which might direct the evidence towards adjacent clauses and slightly reduce the accuracy of the answer, highlighting the importance of improving tolerance for semantic ambiguity by checking the consistency of clauses and conducting more rigorous entity matching.

**5.4 Analysis of typical multimodal cases**

To further examine how the proposed KAG framework improves standard-compliant reasoning, we analyze a representative formula- and table-intensive case concerning the thermal inertia index D of a single homogeneous material layer. This task requires the correct application of normative equations, parameter definitions, and engineering interpretations as specified in GB 50176, rather than relying on general physical intuition.

Baseline models without knowledge augmentation consistently fail on this task. Although their responses are linguistically fluent and technically plausible, they frequently substitute non-normative formulations (e.g., thermal effusivity expressions) for the standard-defined equation $D=R\cdot S$, leading to systematic deviations from regulatory requirements and uniformly low manual scores. This pattern indicates that, in the absence of explicit standard grounding, general-purpose

LLMs tend to prioritize domain-general physics knowledge over code-specific calculation logic.

**Reference answer**

…For a single homogeneous material layer, the thermal inertia index is defined by the code as $D = R \cdot S$ D is dimensionless. R is the layer thermal resistance ($m^2 \cdot K/W$). S is the heat storage coefficient ($W/(m^2 \cdot K)$). …

**base**

…It states $D = \sqrt{(\lambda \rho c)}$ (thermal effusivity), with units W·s^0.5/($m^2 \cdot K$), and defines λ, ρ, c. …

**KAG (our)**

…It follows the code definition and gives $D = R \cdot S$ D is dimensionless, with R and S . It interprets D as the buffering capacity and thermal stability under periodic thermal actions …

**Fig. 19.** Comparison of multi-modal case the results of the base system with KAG system.

In contrast, KAG-enhanced models demonstrate a clear performance improvement, with the largest score gains observed in this category. By explicitly retrieving relevant clauses, formula definitions, and tabulated parameter meanings, the KAG system aligns the generation process with the hierarchical structure of the building standard. As a result, KAG outputs not only reproduce the correct formula, but also present parameter units, engineering meanings, and application contexts consistent with professional design practice.

These results directly reflect the effectiveness of the proposed KAG architecture. The retrieval module constrains the model's reasoning space to authoritative standard knowledge, while the structured fusion mechanism ensures that symbolic expressions and textual explanations remain mutually consistent during generation. This case illustrates that KAG improves reliability not by increasing response length, but by enforcing normative alignment—an essential requirement for intelligent automation in construction engineering.

## 6. Conclusion

Building engineering standards are essential for ensuring safety, quality, and regulatory consistency in construction practice. However, practical standard application is challenged by the massive scale of standard systems, complex cross-document references, heterogeneous knowledge forms, and strong requirements for accuracy and traceability. To address these challenges, this study proposes and implements BEST-KAG, a knowledge-augmented generation framework for intelligent question answering on building engineering standards. The framework integrates unified multimodal knowledge graph modeling, automated knowledge extraction using rule-based parsing and LLM assistance, and knowledge-augmented question answering through graph retrieval combined with LLM generation, enabling clause-level localization and traceable, multimodal-supported standard interpretation. Under consistent inference settings and the proposed 140-question benchmark, we compare mainstream LLM backbones under the text-only baseline and the BEST-KAG-enhanced setting. BLEU-1, BLEU-2, ROUGE-L, cosine similarity and Expert are reported, with Expert as the primary performance indicator. As shown in Fig. 15, BEST-KAG leads to substantial improvements for all compatible models. The Expert score of Claude increases from 0.52 to 0.83, while ChatGPT improves from 0.52 to 0.82. Qwen shows the largest gain, rising from 0.48 to 0.83. In relative terms, the performance improvements reach 58.84%, 57.21%, and 74.01%, respectively, demonstrating that the proposed framework consistently enhances model performance across different LLM backbones. The key contributions of this article can be summarized as follows: (i) A unified multimodal knowledge representation is established to integrate textual clauses, equations, tables, and figures into a traceable knowledge graph while preserving hierarchical and cross-reference structures. (ii) An automated knowledge construction pipeline combining rule-based parsing, knowledge graph modeling, and LLM-assisted semantic analysis is developed to enable scalable extraction of clause structure, cross-clause relations, and multimodal knowledge. (iii) A knowledge-augmented generation framework (BEST-KAG) is developed to support clause-grounded and multimodal evidence-supported question answering for building engineering standards.

Despite the encouraging results, several limitations remain. The current framework is validated on a selected set of building engineering standards, and its generalizability to larger-scale and more heterogeneous standard systems requires further investigation. In addition, the automated knowledge construction still depends on the quality of multimodal content recognition and clause parsing, and the integration with real-world digital design review and automated compliance workflows has not yet been systematically evaluated.

Future research can be advanced along three directions. First, model-adaptive evidence management-such as evidence reranking, structured summarization, and hierarchical context injection-should be explored to mitigate long-context noise and improve cross-model stability. Second, to enable stronger automated compliance capabilities, the clause-parameter-constraint mapping to executable validation code should be strengthened through well-defined intermediate representations and function libraries[38], while explicitly handling unit consistency, boundary-condition recognition, and exception processing. Third, toward Level-4 digitalization, standard knowledge should be more tightly coupled with BIM, point clouds[41], and field data, enabling cross-modal reasoning and closed-loop feedback with the knowledge graph as the hub. With these extensions,[39] BEST-KAG has the potential to evolve into a general-purpose infrastructure for the digital and intelligent use of building standards, supporting governance capacity and high-quality construction in engineering practice.

**Acknowledgments**

The authors are grateful for the financial support received from the National Natural Science Foundation of China (no. 52378306).